\documentclass[letterpaper, 10 pt, conference]{ieeeconf}  

\usepackage{times}
\usepackage{epsfig}
\usepackage{graphicx}
\usepackage{amsmath}
\usepackage{amssymb}

\usepackage{subfigure}
\usepackage{makecell}
\usepackage{multirow}

\usepackage{color}
\usepackage{xcolor}
\usepackage{soul}

\usepackage{cite}
\usepackage{caption}

\IEEEoverridecommandlockouts                              

\title{\LARGE \bf
DVI-SLAM: A Dual Visual Inertial SLAM Network
}

\author{Xiongfeng Peng$^{1}$, Zhihua Liu$^{1}$, Weiming Li$^{1}$, Ping Tan$^{2}$, SoonYong Cho$^{3}$, Qiang Wang$^{1}$
\thanks{Xiongfeng Peng$^{1}$, Zhihua Liu$^{1}$, Weiming Li$^{1}$ and Qiang Wang$^{1}$ are with SAIT-China Lab, Samsung R\&D Institute China-Beijing, China
        {\tt\small \{xf.peng, zhihua.liu, weiming.li, qiang.w\}@samsung.com}}%
\thanks{Ping Tan$^{2}$ with Department of Electronic and Computer Engineering, The Hong Kong University of Science and Technology, China}%
\thanks{SoonYong Cho$^{3}$ with Multimedia System TU, Samsung Advanced Institute of Technology, South Korea}%
}

\begin{document}

\maketitle
\thispagestyle{empty}
\pagestyle{empty}

\begin{abstract}

Recent deep learning based visual simultaneous localization and mapping (SLAM) methods have made significant progress. However, how to make full use of visual information as well as better integrate with inertial measurement unit (IMU) in visual SLAM has potential research value. This paper proposes a novel deep SLAM network with dual visual factors. The basic idea is to integrate both photometric factor and re-projection factor into the end-to-end differentiable structure through multi-factor data association module. We show that the proposed network dynamically learns and adjusts the confidence maps of both visual factors and it can be further extended to include the IMU factors as well. Extensive experiments validate that our proposed method significantly outperforms the state-of-the-art methods on several public datasets, including TartanAir, EuRoC and ETH3D-SLAM. Specifically, when dynamically fusing the three factors together, the absolute trajectory error for both monocular and stereo configurations on EuRoC dataset has reduced by 45.3$\%$ and 36.2$\%$ respectively. 

\end{abstract}

\section{Introduction}


State-of-the-art (SOTA) deep learning-based SLAM methods can be roughly categorized into two classes according to the camera pose estimation process.
The first class consists of direct regression-based methods such as DEMON \cite{ummenhofer2017demon}, DeepTAM \cite{zhou2018deeptam} and DytanVO \cite{shen2023dytanvo}. These methods regress the camera pose with a deep network, which often suffer from limited accuracy and generalization problem.
The other class consists of optimization-based methods such as CodeSLAM \cite{bloesch2018codeslam}, BA-Net \cite{tang2018ba}, DeepFactors \cite{czarnowski2020deepfactors}, and DROID-SLAM \cite{teed2021droid}.
These methods usually include a bundle adjustment (BA) optimizer that minimizes residuals from various related factors corresponding to different aspects of data constraints.
CodeSLAM \cite{bloesch2018codeslam} encodes images into compact representations and infers camera pose by minimizing classical photometric and depth residuals with a non-differentiable damped Gauss-Newton algorithm.
BA-Net \cite{tang2018ba} achieves end-to-end learning by integrating a differentiable optimizer into a deep neural network and infers pose by minimizing photometric residual in feature space (also referred to as feature-metric residual).
However, these photometric-based methods typically need a good initial guess for the optimizer to work well.
DeepFactors \cite{czarnowski2020deepfactors} proposes to combine photometric, re-projection, and depth factors on CodeSLAM \cite{bloesch2018codeslam}, but without considering each factor's reliability.
Later, DROID-SLAM \cite{teed2021droid} learns to optimize camera pose and depth iteratively by minimizing the re-projection residual through a dense bundle adjustment (DBA) optimizer.
However, the re-projection-based method heavily relies on correct data association, and noisy correlations could cause severe errors in the estimated flow and result in inaccurate camera pose.
The above mentioned methods use only visual factors.
It is well-known that including an additional inertial measurement unit (IMU) can enhance the robustness of visual SLAM methods, especially in fast movement. Some direct regression-based deep-learning SLAM methods have tried to integrate IMU factors such as VINet \cite{clark2017vinet} and DeepVIO \cite{han2019deepvio}.
But how to effectively fuse all the different visual factors and IMU factors into an end-to-end deep network still poses an open question for the research community.

In this paper, we propose a novel optimization-based deep SLAM method, which is named Dual Visual Inertial SLAM network (DVI-SLAM). DVI-SLAM simultaneously infers camera pose and dense depth by dynamically fusing multiple factors with an end-to-end trainable differentiable structure. Here, the dual visual factors refer to the feature-metric factor and the re-projection factor, which provide complimentary cues to explore the visual information. The network also extends naturally to effectively include the IMU factor or other factors.
A key design of our DVI-SLAM is its dynamic multi-factor fusion that learns confidence maps to adjust relative strengths among the different factors for each iteration in the optimization process.
In particular, the re-projection factor dominates at the early stage of BA optimization to get a good initialization and avoid getting stuck in local minima.
In the later stage, both the feature-metric factor and re-projection factor smoothly steer the joint optimization towards a true minimum. 
The network also dynamically adjusts the confidence map of IMU factor according to the estimated pose error for a reliable visual-inertial fusion. 
Our dual visual and IMU constraints are fused and minimized in a tightly-coupled manner. To the best of our knowledge, our DVI-SLAM is the first deep learning-based framework that allows tightly coupling dual visual inertial SLAM with DBA optimization.


Our main contributions are summarized as follows:
\begin{itemize}
    \item We propose a DVI-SLAM network that dynamically fuses re-projection, feature-metric, and IMU factors with learned confidence maps to improve camera pose estimation in challenging scenes.
    \item Our proposed end-to-end learnable framework is flexible to include various factors, supporting monocular or stereo, with or without depth or IMU sensors.
    \item Our method exceeds the SOTA methods on several public datasets: TartanAir, EuRoC and ETH3D-SLAM.
    \end{itemize}

\section{Related Work}
The existing SLAM methods are categorized according to the factors in the optimization process and they are introduced as follows.

\begin{figure*}[t]
    \begin{center}
    \includegraphics[width=0.9\linewidth]{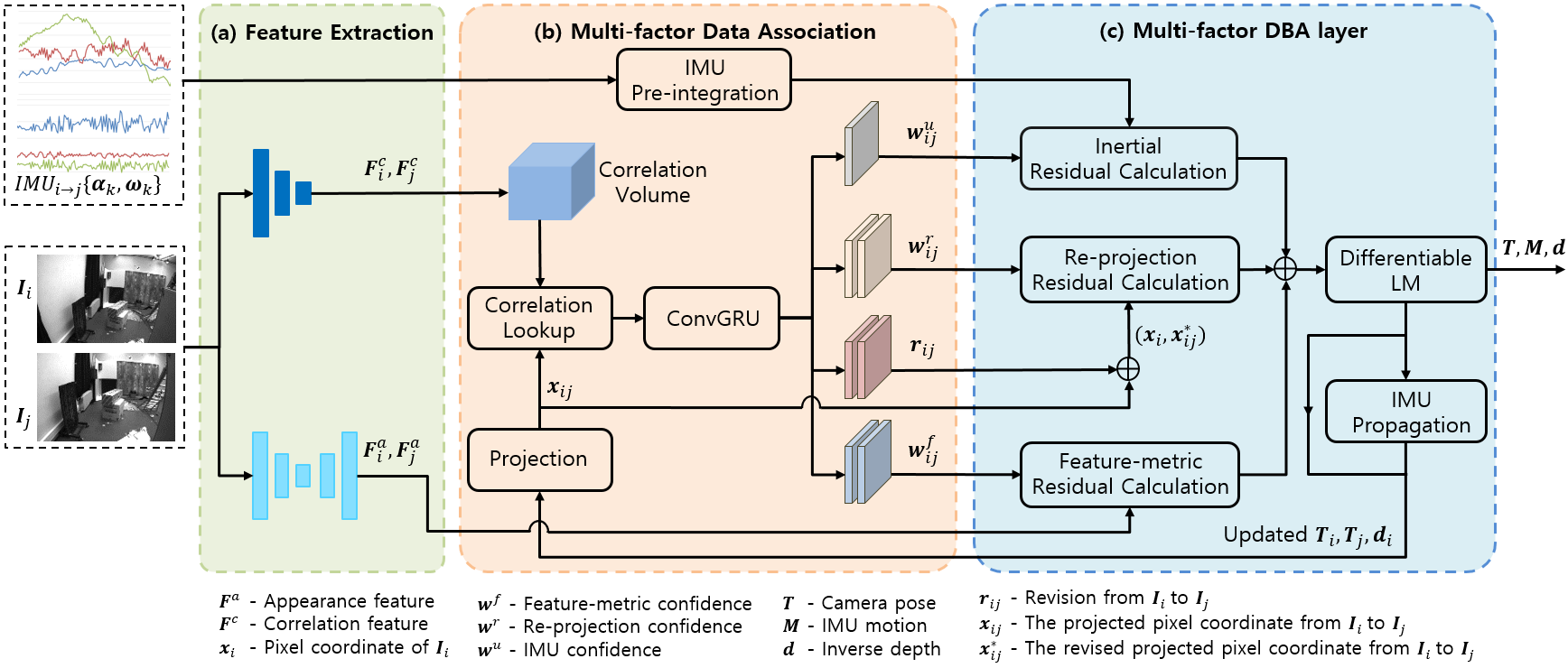}
    \end{center}
    \caption{Overview of our DVI-SLAM structure (the two-views reconstruction only for better clarity). (a) Visual correlation feature and appearance feature are extracted by the input two frames $\boldsymbol{I}_{i}, \boldsymbol{I}_{j}$.
    (b) The multi-factor data association module learns the feature correspondence and confidence maps, and IMU pre-integration results are calculated by the accelerometer and gyroscope measurements of IMU $\{\boldsymbol{\alpha}_{k}, \boldsymbol{\omega}_{k}\}$ between the two frames.
    (c) Those confidence maps dynamically fuse re-projection, feature-metric, and IMU residuals for optimization via a differentiable LM for the pose $\boldsymbol{T}$, IMU motion $\boldsymbol{M}$, and inverse depth $\boldsymbol{d}$. $\boldsymbol{T}$ is also calculated via the IMU propagation.
    Finally, $\boldsymbol{T}$, $\boldsymbol{M}$, and $\boldsymbol{d}$ are updated for the next iteration.
    }
    \label{fig-network}
    \end{figure*}

\paragraph{Re-projection-based SLAM Methods}
These methods first extract keypoints and descriptors from images and then build data association \cite{davison2007monoslam, klein2007parallel, mourikis2007multi, leutenegger2015keyframe, eckenhoff2019closed, campos2021orb,liu2018ice,leutenegger2015keyframe,qin2018vins}. The estimated camera pose and 3D map are optimized with BA by minimizing the re-projection error. PTAM \cite{klein2007parallel} is the first real-time optimization-based SLAM system in this category.
Later, ORB-SLAM3 \cite{campos2021orb} combines the re-projection factors and fuses it with an IMU sensor on the basis of \cite{mur2015orb, mur2017orb, mur2017visual}. Recently, DROID-SLAM \cite{teed2021droid} proposes a deep dense feature-based SLAM network to iteratively optimize the camera pose and dense depth.
Based on DROID-SLAM, DPVO \cite{teed2022deep} proposes a deep visual odometry system built using a sparse patch representation, and PVO \cite{ye2023pvo} proposes a panoptic visual odometry framework to achieve more comprehensive modeling of the scene motion, geometry, and panoptic segmentation information.

\paragraph{Photometric-based SLAM Methods}
Photometric-based SLAM methods \cite{forster2014svo, eckenhoff2017direct, engel2014lsd, engel2017direct, engel2015large, wang2017stereo} generally utilize raw pixel intensities and define an objective function with photometric error. Similar to photometric factors, feature-metric factors use deep features extracted by neural networks instead of pixel intensities.
CodeSLAM \cite{bloesch2018codeslam} encodes observed images into a compact and optimizable representation that is used in a keyframe-based SLAM system to estimate both camera poses and depths.
BA-Net \cite{tang2018ba} minimizes the feature-metric residual with the combined learned depth basis, which has poor generalization and does not converge well.
One advantage of photometric-based SLAM methods is that they allow the inclusion of abundant image information. This empowers these methods to work well as long as the projection residual is reasonably small instead of requiring correct explicit data association.

\paragraph{IMU-related SLAM Methods}
IMU sensor is widely used in traditional SLAM systems such as ORB-SLAM3 \cite{campos2021orb}, ICE-BA \cite{liu2018ice}, VINS \cite{qin2018vins}, and OKVIS \cite{leutenegger2015keyframe}.
For deep learning-based SLAM methods, there are several IMU-related works. VINet \cite{clark2017vinet} is the first deep visual-inertial SLAM network that directly concatenates the deep features with IMU feature and directly regresses the camera pose.
DeepVIO \cite{han2019deepvio} fuses the visual and inertial features with a designed fully connected layer without considering the reliability of different sensors.
Closely related to the above three categories of methods, our proposed DVI-SLAM method dynamically fuses re-projection, feature-metric, and IMU factors in an end-to-end differentiable network.

\section{Two-view Reconstruction}

In this section, we introduce our DVI-SLAM with two-view reconstruction for clarity. The complete system is described in Section~\ref{sec:system}. Here, the two-view reconstruction refers to the estimation of camera pose, IMU motion and inverse depth from two views.

Figure \ref{fig-network} illustrates our DVI-SLAM network for the two-view reconstruction.
Given inputs of images $\boldsymbol{I}_{i}$, $\boldsymbol{I}_{j}$ and the accelerometer and gyroscope measurements of IMU $\{\boldsymbol{\alpha}_{k}, \boldsymbol{\omega}_{k}\}$  between the two frames, the network outputs the updated camera pose $\boldsymbol{T}_{i}$,$\boldsymbol{T}_{j}$, IMU motion $\boldsymbol{M}_{i}$,$\boldsymbol{M}_{j}$ and dense inverse depth $\boldsymbol{d}_{i}$.
Where the camera pose $\boldsymbol{T}=(\boldsymbol{R}, \boldsymbol{p})$ contains rotation $\boldsymbol{R}$ and translation $\boldsymbol{p}$. The IMU motion $\boldsymbol{M}=(\boldsymbol{v}, \boldsymbol{ba}, \boldsymbol{bg})$ includes velocity $\boldsymbol{v}$, accelerator bias $\boldsymbol{ba}$, and gyroscope bias $\boldsymbol{bg}$. 
Our DVI-SLAM network is built upon DROID-SLAM\cite{teed2021droid}. It consists of two modules and one layer: (a) a feature extraction module, (b) a multi-factor data association module, and (c) a multi-factor DBA layer.
In the following sections, we introduce these modules and layers respectively.

\subsection{Feature Extraction}

In the feature extraction module, two kinds of features are extracted. One feature is the correlation feature $\boldsymbol{F}^{c}$ which is used to estimate a dense flow map and build the correspondence between two frames. The other is the deep appearance feature $\boldsymbol{F}^{a}$ with more details for computing the feature-metric residual.

Different from DROID-SLAM\cite{teed2021droid}, we add an encoder-decoder branch with a U-Net structure \cite{ronneberger2015u} to extract the appearance feature. The encoder has two sub-modules and each contains two convolutional layers followed by a downsampling layer and max pooling operator. The decoder also has two sub-modules and the feature map resolution increases through sequential upsampling until matches the resolution of the input image.

For the module outputs, the resolution of the appearance feature map is eight times that of the correlation feature map. More details are kept in the deep appearance feature map by the skip connection of the U-Net structure.

\subsection{Multi-factor Data Association}

The multi-factor data association module learns the feature correspondence and the confidence maps for fusion with re-projection, feature-metric, and IMU.
Figure~\ref{fig-network} (b) illustrates the detailed process of one iteration.
Similar to DROID-SLAM \cite{teed2021droid}, a pyramid 4D correlation volume $\boldsymbol{C}_{ij}$ is built by inputs from the extracted correlation features $\boldsymbol{F}_{i}^{c}, \boldsymbol{F}_{j}^{c}$. 
And the projected pixel coordinate $\boldsymbol{x}_{ij}$ is calculated by the updated pose $\boldsymbol{T}_{i}, \boldsymbol{T}_{j}$, inverse depth $\boldsymbol{d}_{i}$ from last iteration.
Then the network looks up neighbourhood correlation feature maps from the 4D correlation volume $\boldsymbol{C}_{ij}$ with $\boldsymbol{x}_{ij}$ and feed the correlation feature maps into a ConvGRU network.
The $\boldsymbol{x}_{ij}$ represents the projected pixel coordinates from the pixel coordinates $\boldsymbol{x}_{i}$ of $\boldsymbol{I}_i$ to $\boldsymbol{I}_j$ using pose $\boldsymbol{T}_{i}, \boldsymbol{T}_{j}$, inverse depth $\boldsymbol{d}_{i}$.
Specifically, $\boldsymbol{x}_{ij}=\Pi(\boldsymbol{T}_{i}, \boldsymbol{T}_{j}, \boldsymbol{d}_{i}, \boldsymbol{x}_{i})$, where $\Pi$ represents the projection.

Since the deep network has the capability to learn complex priors, we design a multi-head GRU module to learn the uncertainty of different factors for subsequent dynamic residual fusion.
Specifically, the multi-head GRU module has four independent output branches: IMU confidence branch, re-projection confidence branch, feature-metric confidence branch, and flow revision branch.
For the IMU confidence branch, it has one $1\times1$ convolution and one average pooling layer.
For the other three branches, two $3\times3$ convolutional layers are applied. In addition, to predict the confidence maps of re-projection and feature-metric branches, one sigmoid operator is used for each branch respectively.

\begin{figure}[t]
    \begin{center}
    \includegraphics[width=0.95\linewidth]{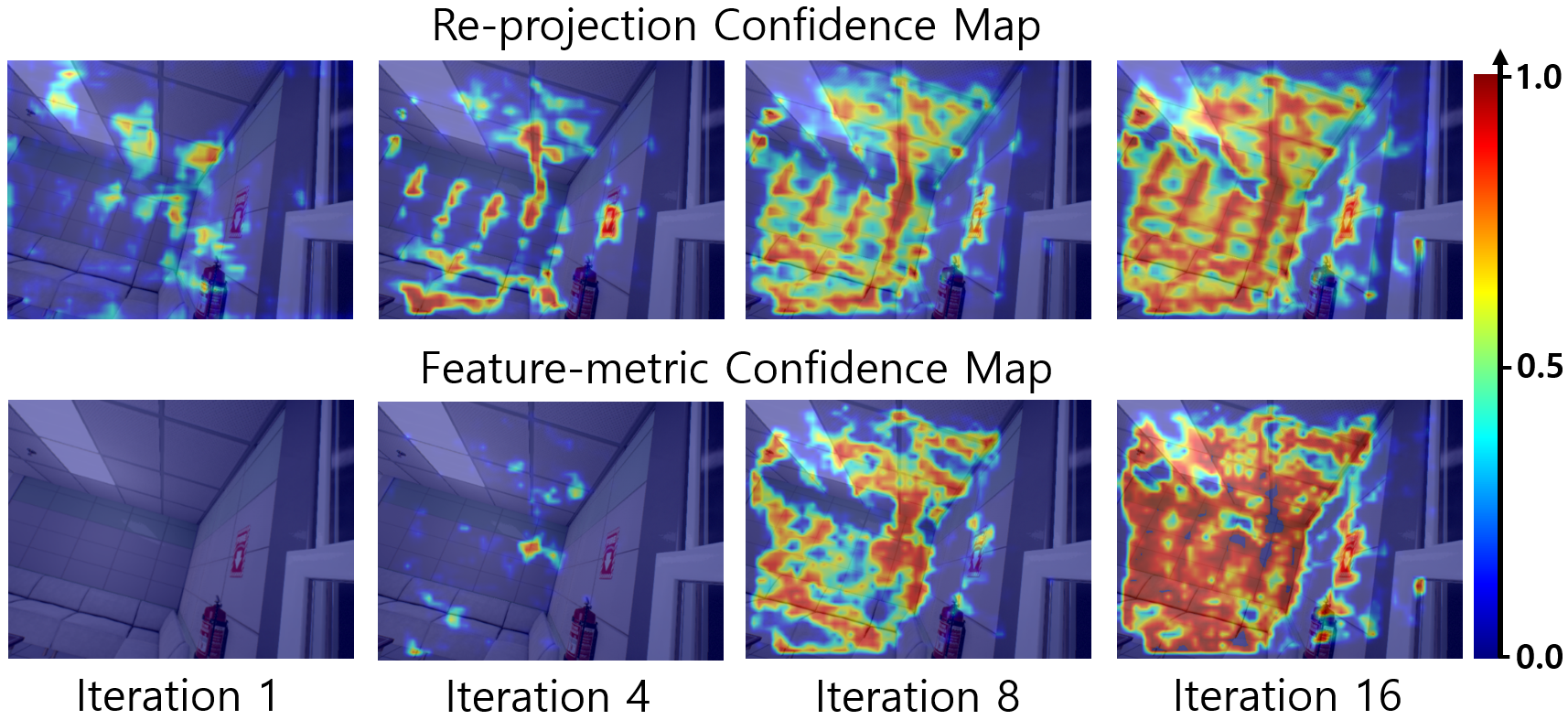}
    \end{center}
    \caption{The confidence map changes iteratively. High confidence in the re-projection confidence map focuses on corner pixels and extends to edge pixels gradually with iteration increasing. For the feature-metric confidence map, the confidence is zero at the beginning of the iteration and focuses on the well-textured regions after several iterations.}
    \label{fig_confidence}
    \end{figure}

Figure \ref{fig_confidence} visualizes our learned re-projection and feature-metric confidence maps. We notice that at the beginning of the optimization iteration, the confidence map of the feature-metric factor is small and the re-projection factor is large. When the iteration continues, the confidence map of feature-metric factor increases gradually, and it will be slightly larger than the re-projection factor at the end of the iteration.
In both re-projection and feature-metric confidence maps, the locations with high confidence are well-textured. It's reasonable because the well-textured areas easily build explicit and implicit correspondences. 
The multi-head GRU module learns IMU confidence as well. The insight is that the pose from the DBA layer is updated again through IMU propagation.
In this step, the IMU information is implicitly injected into the multi-head GRU module in the network.
Based on the learned confidence, Mahalanobis distance is used to weigh the residual of each factor. The larger the average confidence, the more reliable the factor.

\subsection{Multi-factor DBA Layer}
With the multi-factor confidences, the residuals are fused and updated iteratively in the DBA layer.
The following section introduces the confidence-based residual definition of each factor in detail.

\paragraph{Confidence-based Re-projection Residual}
Same as DROID-SLAM \cite{teed2021droid}, according to the correspondence $(\boldsymbol{x}_{i}, \boldsymbol{x}^{*}_{ij})$ of the pixel coordinates and the learned re-projection confidence map $\boldsymbol{w}_{ij}^r$, the re-projection residual for two-view is defined as in Equation~\ref{equ:VRC}.

\begin{footnotesize}
\begin{equation}\label{equ:VRC}
    \begin{aligned}
        E^{r}(\boldsymbol{T}, \boldsymbol{d})\!=\!||\boldsymbol{x}^{*}_{ij}\!-\!\Pi(\boldsymbol{T}\!_{i}, \boldsymbol{T}\!_{j}, \boldsymbol{d}_{i}, \boldsymbol{x}_{i})||_{\sum_{ij}^r}^{2}, \quad {\sum\nolimits_{ij}^r\!\!=\!\!diag \boldsymbol{w}_{ij}^r}
    \end{aligned}
    \end{equation}
\end{footnotesize}, where $||\cdot||_{\sum_{ij}}$ is the Mahalanobis distance.
The larger the confidences, the more reliable the correspondences.
The residual of each correspondence in the re-projection term updates iteratively.

\paragraph{Confidence-based Feature-metric Residual}
Based on the appearance features $\boldsymbol{F}_{i}^{a}, \boldsymbol{F}_{j}^{a}$ and the learned feature-metric confidence $\boldsymbol{w}_{ij}^f$, we calculate the two-view feature-metric residual as in Equation~\ref{equ:VFC}.

\begin{footnotesize}
\begin{equation}\label{equ:VFC}
    \begin{aligned}
        E^{f}(\boldsymbol{T}, \boldsymbol{d})\!=\!||\boldsymbol{F}\!_{i}^{a}(\boldsymbol{x}_{i})\!\!-\!\!\boldsymbol{F}\!_{j}^{a}(\Pi(\boldsymbol{T}\!_{i}, \boldsymbol{T}\!_{j}, \boldsymbol{d}_{i}, \boldsymbol{x}_{i}))||_{\sum_{ij}^f}^{2}, \quad \!\!\!{\sum\nolimits_{ij}^f\!\!=\!\!diag \boldsymbol{w}_{ij}^f}
    \end{aligned}
    \end{equation}
\end{footnotesize}

\paragraph{Confidence-based Inertial Residual}
Following the traditional VI-SLAM methods \cite{qin2018vins, liu2018ice}, we compute the inertial residual between image $\boldsymbol{I}_i$ and $\boldsymbol{I}_j$.
Based on IMU pre-integration \cite{forster2016manifold}, the $\Delta$ in Equation~\ref{equ:ICT} and Equation~\ref{equ:ICM}, and the pre-integration weight $\boldsymbol{w}_{ij}^{pu}$ can be estimated.
The bias $\boldsymbol{\hat{b}}_i=(\boldsymbol{\hat{ba}}_i, \boldsymbol{\hat{bg}}_i)$ is the bias estimate at the time of pre-integration.
The IMU weight $\boldsymbol{w}_{ij}^{imu}$ is represented by $\boldsymbol{w}_{ij}^{imu}=\boldsymbol{w}_{ij}^{pu}\boldsymbol{w}_{ij}^{u}$, where $\boldsymbol{w}_{ij}^{u}$ is the learned IMU confidence.
It should be noted that there is an extrinsic difference between the IMU pose and the camera pose.
We calculate the two-view inertial residual as in Equation~\ref{equ:IC}.

\begin{footnotesize}
\begin{equation}\label{equ:IC}
    \begin{aligned}
    E^{u}(\boldsymbol{T}, \boldsymbol{M})=\left|\left|\!\!\!
    \begin{array}{c}
    e_{ij}^{T}\\
    e_{ij}^{M}\\
    \end{array}
\!\!\!\right|\right|_{\sum_{ij}^u}^{2},  \quad {\sum\nolimits_{ij}^u=diag \boldsymbol{w}_{ij}^{imu}}
\end{aligned}
\end{equation}
\end{footnotesize}

\begin{footnotesize}
\begin{equation}\label{equ:ICT}
    \begin{aligned}
    e_{ij}^{T}\!=\!\!
\left[\!\!\!\!
    \begin{array}{c}
    Log((\Delta \boldsymbol{R}_{ij}Exp(\Delta \boldsymbol{J}_{ij}^r(\boldsymbol{bg}_i\!-\!\boldsymbol{\hat{bg}}_i)))^T \boldsymbol{R}_i^T \boldsymbol{R}_j)\\
    \boldsymbol{R}_i^T(\boldsymbol{p}_j\!-\!\boldsymbol{p}_i\!-\!\boldsymbol{v}_i \Delta t_{ij}\!-\!\frac{1}{2} \boldsymbol{g} t_{ij}^2)\!-\!(\Delta \boldsymbol{p}_{ij}\!+\!\Delta \boldsymbol{J}_{ij}^p(\boldsymbol{b}_i\!-\!\boldsymbol{\hat{b}}_i))\\
    \end{array}
\!\!\!\!\right]
\end{aligned}
\end{equation}
\end{footnotesize}

\begin{footnotesize}
\begin{equation}\label{equ:ICM}
    \begin{aligned}
    e_{ij}^{M}=
\left[\!\!\!\!
    \begin{array}{c}
    \boldsymbol{R}_i^T(\boldsymbol{v}_j\!-\!\boldsymbol{v}_i\!-\!\boldsymbol{g} t_{ij})\!-\!(\Delta \boldsymbol{v}_{ij}\!+\!\Delta \boldsymbol{J}_{ij}^v(\boldsymbol{b}_i\!-\!\boldsymbol{\hat{b}}_i))\\
    \boldsymbol{ba_j}-\boldsymbol{ba_i}\\
    \boldsymbol{bg_j}-\boldsymbol{bg_i}\\
    \end{array}
\!\!\!\!\right]
\end{aligned}
\end{equation}
\end{footnotesize}

\paragraph{Multi-factor Residual Fusion}
After getting the re-projection, feature-metric, and IMU residuals of two-view, the total residual $E(\boldsymbol{T}, \boldsymbol{M}, \boldsymbol{d})$ in Equation~\ref{equ:DVIC-two-view} is obtained by summing the residuals of each factor and is minimized in the DBA layer. After multiple iterations and updates, the system outputs the optimal camera pose $\boldsymbol{T}$, inverse depth $\boldsymbol{d}$, and IMU motion $\boldsymbol{M}$.

\begin{footnotesize}
\begin{equation}\label{equ:DVIC-two-view}
    \begin{aligned}
    E(\boldsymbol{T}, \boldsymbol{M}, \boldsymbol{d})=E^{r}(\boldsymbol{T}, \boldsymbol{d})+E^{f}(\boldsymbol{T}, \boldsymbol{d})+E^{u}(\boldsymbol{T}, \boldsymbol{M})
    \end{aligned}
    \end{equation}
\end{footnotesize}

\section{DVI-SLAM System}
\label{sec:system}

Same as DROID-SLAM \cite{teed2021droid}, for an input video stream, multiple keyframes are extracted to construct a frame graph $G = (\boldsymbol{V},\boldsymbol{\varepsilon})$ which indicates the co-visibility among these keyframes. 
Equation~\ref{equ:DVIC-two-view} is extended into Equation~\ref{equ:DVIC-framegraph} and optimizes the total residual of the entire frame graph via a differentiable Levenberg-Marquardt (LM).

\begin{footnotesize}
\begin{equation}\label{equ:DVIC-framegraph}
    \begin{aligned}
    E(\boldsymbol{T}, \boldsymbol{M}, \boldsymbol{d})=\!\!\!\!\sum_{(i,j) \in \boldsymbol{\varepsilon}}\!\!\!\!E^{r}(\boldsymbol{T}, \boldsymbol{d})+\!\!\!\!\sum_{(i,j) \in \boldsymbol{\varepsilon}}\!\!\!\!E^{f}(\boldsymbol{T}, \boldsymbol{d})+\!\!\!\!\!\!\sum_{(i,j=i+1) \in \boldsymbol{\varepsilon}}\!\!\!\!\!\! E^{u}(\boldsymbol{T}, \boldsymbol{M})
    \end{aligned}
    \end{equation}
\end{footnotesize}

In the following section, we introduce the DVI-SLAM system initialization and the process in detail.

\paragraph{Initialization}
Same as DROID-SLAM method \cite{teed2021droid}, we initialize DVI-SLAM when there is only visual stream input.
When there is both visual and inertial stream inputs, we follow VINS-Mono method \cite{qin2018vins} for DVI-SLAM initialization. 
Specifically, there are three steps for initialization.
Firstly, multiple keyframes are collected to build a frame graph, and camera pose and inverse depth states are optimized by DBA layer. Then we initialize the gravity direction and IMU motion according to the estimated camera pose and IMU pre-integration result.
For the monocular visual and inertial streams, the absolute scale is also initialized.
Finally, we minimize re-projection residual, feature-metric residual, and inertial residual with DBA layer to refine camera pose, motion, and inverse depth.

\paragraph{VI-SLAM process}
After the system initialization is finished, for a new frame and its corresponding IMU data, we first estimate the average ﬂow between the new frame and its latest keyframe. If the new frame is considered a keyframe, it is added to the sliding window and the frame graph which establishes edges with its temporally adjacent keyframes. 
The new frame pose and IMU motion are optimized in the sliding window with local BA.
We use the same strategy as DROID-SLAM \cite{teed2021droid} to keep a constant sliding window size and get the poses of non-keyframes. 
The camera pose after optimizing with local BA is called visual-inertial odometry (VIO) output. Based on the VIO result in the sequence, we build a new frame graph in which the keyframes are connected spatially and temporally. After performing an extra global BA, the results are called VI-SLAM output which reﬂects the accuracy of the reconstructed global map.

\section{Experiments}

To validate our DVI-SLAM system, we make extensive experiments on both synthetic and real-world datasets with different sensor combinations.
Specifically, we use public TartanAir dataset \cite{wang2020tartanair}, TartanAir SLAM challenge \cite{wang2020tartanair}, EuRoC dataset \cite{burri2016euroc}, and ETH3D-SLAM benchmark \cite{schops2019bad}. 
The absolute trajectory error (ATE) \cite{sturm2012benchmark} metric
is used to evaluate the error of the pose.
Since there is no absolute scale for monocular input, we additionally optimize the scale (Sim3) when evaluating the trajectory error and name the metric as ATE-scale.
In order to better illustrate the effectiveness and robustness of our method, we compare our DVI-SLAM with both traditional and learning-based visual/visual-inertial odometry (VO/VIO) and visual/visual-inertial SLAM (V/VI-SLAM) systems.
Due to limited space, we provide more experimental results in the supplementary material.

\begin{table*}[t]
    \begin{center}
    \caption{The trajectory error comparison of our method with SOTA methods on TartanAir monocular SLAM challenge.}
    \label{TartanAir_M_challenge}
    \resizebox{16cm}{!}{
    \begin{tabular}{ll|cccccccc|c}\hline
    \multicolumn{2}{c|}{Mono(ATE-scale(m))$\downarrow$} &MH000 &MH001 &MH002 &MH003 &MH004 &MH005 &MH006 &MH007 &Avg \\ \hline
    \multirowcell{4}{VO}  &DeepV2D\cite{teed2018deepv2d} &6.15 &2.12 &4.54 &3.89 &2.71 &11.55 &5.53 &3.76 &5.03\\
    &TartanVO\cite{wang2021tartanvo} &4.88 &0.26 &2.00 &0.94 &1.07 &3.19 &1.00 &2.04 &1.92\\
    &DROID-SLAM*\cite{teed2021droid}  &0.50 &0.08 &0.10 &0.06 &1.33 &0.53 &0.13 &2.18 &0.61\\
    &DVI-SLAM(Ours) &\textbf{0.19} &\textbf{0.04} &\textbf{0.04} &\textbf{0.03} &\textbf{0.93} &\textbf{0.11} &\textbf{0.08} &\textbf{1.19} &\textbf{0.33}\\ \hline
    \multirowcell{4}{SLAM}  &ORB-SLAM\cite{mur2015orb} &1.30 &0.04 &2.37 &2.45 &- &- &21.47 &2.73 &-\\
     &DROID-SLAM\cite{teed2021droid} &\textbf{0.08} &0.05 &0.04 &0.02 &\textbf{0.01} &1.31 &0.30 &\textbf{0.07} &0.24\\
    &DROID-SLAM*\cite{teed2021droid} &0.58 &0.05 &0.03 &0.03 &1.30 &0.52 &0.13 &0.14 &0.35\\
    &DVI-SLAM(Ours) &\textbf{0.08} &\textbf{0.03} &\textbf{0.02} &\textbf{0.01} &0.63 &\textbf{0.12} &\textbf{0.06} &0.09 &\textbf{0.13}\\ \hline
    \end{tabular}
    }
    \end{center}
    \end{table*}

\begin{table*}[t]
    \begin{center}
    \caption{The trajectory error comparison of our method with the SOTA VO and VIO methods on EuRoC dataset.}
    \label{tab_EuRoc_VO_SOTA}
    \resizebox{16cm}{!}{
    \begin{tabular}{ll|ccccc|ccc|ccc|c}\hline
    \multicolumn{2}{c|}{VO/VIO} &MH01 &MH02 &MH03 &MH04 &MH05 &V101 &V102 &V103 &V201 &V202 &V203 &Avg \\ \hline
    \multirowcell{8}{Mono\\ATE-scale(m)$\downarrow$}  &DSO\cite{engel2017direct}  &\textbf{0.046} &\textbf{0.046} &0.172 &3.810 &\textbf{0.110}  &0.089 &\textbf{0.107} &0.903   &\textbf{0.044} &0.132 &1.152  &0.601 \\
    &SVO\cite{forster2016svo}  &0.100 &0.120 &0.410 &0.430 &0.300   &0.070 &0.210 &-   &0.110 &0.110 &1.080  &- \\
    &DeepV2D\cite{teed2018deepv2d}  &0.739 &1.144 &0.752 &1.492 &1.567   &0.981 &0.801 &1.570   &0.290 &2.202 &2.743  &1.298 \\
    &TartanVO\cite{wang2021tartanvo}  &0.639 &0.325 &0.550 &1.153 &1.021   &0.447 &0.389 &0.622   &0.433 &0.749 &1.152  &0.680 \\
    &DROID-SLAM\cite{teed2021droid}  &0.163 &0.121 &0.242 &0.399 &0.270   &0.103 &0.165 &0.158   &0.102 &0.115 &\textbf{0.204}  &0.186 \\
    &DROID-SLAM*\cite{teed2021droid}  &0.063 &0.156 &0.207 &0.264 &0.495   &0.092 &0.204 &\textbf{0.087}   &0.116 &0.080 &0.207  &0.179 \\
    &DVI-SLAM(Ours)  &0.120 &0.063 &\textbf{0.076} &\textbf{0.161} &0.354   &\textbf{0.060} &0.207 &0.113   &0.087 &\textbf{0.061} &0.268  &\textbf{0.143} \\ \hline

    \multirowcell{4}{Mono+IMU\\ATE(m)$\downarrow$}
    &MSCKF\cite{mourikis2007multi}  &0.420 &0.450 &0.230 &0.370 &0.480  &0.340 &0.200 &0.670   &0.100 &0.160 &1.130 &0.414 \\
    &OKVIS\cite{leutenegger2015keyframe}  &0.160 &0.220 &0.240 &0.340 &0.470   &0.090 &0.200 &0.240   &0.130 &0.160 &0.290  &0.231 \\
    &VINS-Mono\cite{qin2018vins}  &0.270 &0.120 &0.130 &0.230 &0.350   &\textbf{0.070} &\textbf{0.100} &0.130   &\textbf{0.080} &0.080 &0.210  &0.161 \\
    &DVI-SLAM(Ours)  &\textbf{0.063} &\textbf{0.083} &\textbf{0.101} &\textbf{0.187} &\textbf{0.163}   &0.074 &0.114 &\textbf{0.083}   &0.091 &\textbf{0.045} &\textbf{0.072}  &\textbf{0.098} \\ \hline

    \multirowcell{5}{Stereo\\ATE(m)$\downarrow$}
    &SVO\cite{forster2016svo}  &\textbf{0.040} &0.070 &0.270 &0.170 &0.120   &\textbf{0.040} &\textbf{0.040} &0.070   &0.050 &0.090 &0.790  &0.159 \\
    &VINS-Fusion\cite{qin2019general}  &0.540 &0.460 &0.330 &0.780 &0.500   &0.550 &0.230 &-   &0.230 &0.200 &-  &- \\
    &DROID-SLAM*\cite{teed2021droid}  &0.065 &0.042 &0.098 &0.191 &0.133   &0.063 &0.045 &0.043   &\textbf{0.040} &0.054 &0.101  &0.080 \\
    &DVI-SLAM(Ours)  &0.043 &\textbf{0.041} &\textbf{0.064} &\textbf{0.148} &\textbf{0.114}   &0.063 &\textbf{0.040} &\textbf{0.031}   &0.049 &\textbf{0.050} &\textbf{0.080}  &\textbf{0.066} \\ \hline

    \multirowcell{4}{Stereo+IMU\\ATE(m)$\downarrow$}
    &OKVIS\cite{leutenegger2015keyframe}  &0.188 &0.123 &0.192 &0.185 &0.292   &\textbf{0.045} &0.088 &0.123   &0.056 &0.114 &0.199  &0.146 \\
    &VINS-Fusion\cite{qin2019general}  &0.159 &0.156 &0.126 &0.279 &0.290   &0.078 &0.068 &0.112   &0.059 &0.091 &0.096  &0.138 \\
    &OKVIS2\cite{leutenegger2022okvis2}  &0.049 &\textbf{0.040} &0.084 &0.115 &0.156   &0.047 &0.075 &0.037  &0.042 &0.042 &\textbf{0.047}  &0.076 \\
    &DVI-SLAM(Ours)  &\textbf{0.042} &0.046 &\textbf{0.081} &\textbf{0.072} &\textbf{0.069}   &0.059 &\textbf{0.034} &\textbf{0.028}   &\textbf{0.040} &\textbf{0.039} &0.055  &\textbf{0.051} \\ \hline
    \end{tabular}
    }
    \end{center}
    \end{table*}

\subsection{Training and Implementation}

\paragraph{Training}
We train the network with a weighted sum of pose loss $L_{pose}$, flow loss $L_{flow}$, re-projection loss $L_{repro}$, and feature-metric loss $L_{metric}$.
Following DROID-SLAM \cite{teed2021droid}, the pose loss $L_{pose}$ is the distance between ground truth (GT) pose and the predicted pose, and the $L_{flow}$ flow loss is the distance between the predicted flow and GT flow which is converted from GT pose and depth.
Re-projection loss $L_{repro}$ and feature-metric loss $L_{metric}$ are the distances of the re-projection error and feature-metric error.
The total training loss $L$ is expressed in Equation~\ref{equ:total_loss} and the hyperparameters $w1, w2, w3, w4$ are set $10, 0.05, 0.01, 0.01$ respectively:

\begin{small}
\begin{equation}\label{equ:total_loss}
    \begin{aligned}
    L = w1\!*\!L_{pose} \!+\! w2\!*\!L_{flow} \!+\! w3\!*\!L_{repro} \!+\! w4 * L_{metric}
    \end{aligned}
    \end{equation}
\end{small}

\paragraph{Implementation}
We train our DVI-SLAM network on four NVIDIA RTX 3090 GPUs with three stages.
In the first two stages, our network is trained on monocular video stream with $384 \times 512$ resolution from the TartanAir dataset \cite{wang2020tartanair}. Each training example consists of a 7-frame video sequence. We fix the first two poses to GT poses of each training sequence to remove the 6-DOF and scale gauge freedoms for learning visual-related parameters.

In the first stage, similar to DROID-SLAM \cite{teed2021droid}, the network learns correlation feature module, ConvGRU module, re-projection confidence branch and flow revision branch and is trained for 250k steps.
The network is supervised with GT pose, GT flow, and re-projection consistency.

In the second stage, we first freeze the parameters learned from the last stage. Then we learn appearance feature module and feature-metric confidence branch for 50k steps. The network is supervised by GT pose, GT flow, and re-projection consistency for the first 10k steps, and then is jointly supervised with feature-metric consistency for the remaining 40k steps.

In the third stage, the network learns IMU confidence branch. We train it on monocular video stream with $384 \times 512$ resolution and IMU stream from the ETH3D-SLAM dataset \cite{schops2019bad}.
We input GT pose and motion of the first frame, without fixing any frame.
Since the ETH3D-SLAM dataset has only GT pose annotations, the average flow cannot be calculated, so we dynamically generate videos by sampling paths in the number of frame intervals between $5$ and $10$.
In this stage, we first freeze the parameters learned from the last two stages, then the IMU confidence branch is learned for 10k steps and is supervised by GT pose, re-projection, and feature-metric consistency.

For all the above three-stage training processes, the batch size, frame clips, and data association update iterations are set to 4, 7, and 15 respectively.

\subsection{Comparison with the SOTA Methods}

In this section, we compare the VO/VIO and V/VI-SLAM output of our proposed DVI-SLAM method with the SOTA methods.
For a fair comparison, we train DROID-SLAM \cite{teed2021droid} from scratch. Our trained DROID-SLAM model has nearly the same accuracy as the author claimed and is considered as the baseline (DROID-SLAM$^*$).


\paragraph{Results on Synthetic Data}
Table \ref{TartanAir_M_challenge} shows the trajectory error comparison of our DVI-SLAM with the SOTA methods on the TartanAir monocular SLAM challenge.
We compare the trajectory error across all Hard sequences.
From the tables, we can see that the results of our method are significantly better than the SOTA methods except for the MH004 sequence.
After examining the sequence, we found that a few frames with mostly a white wall lead to larger pose estimation errors.




\paragraph{Results on Real-world Data}
Besides the synthetic dataset, we also compare our DVI-SLAM with the SOTA methods on real-world visual-inertial datasets.
Table \ref{tab_EuRoc_VO_SOTA} compares the trajectory error of our method with the SOTA VO and VIO methods on EuRoC dataset.
As shown in the table, our VO result exceeds the DROID-SLAM's result for most of the sequences.
In comparison with other SOTA traditional or learning-based VO and VIO methods, our method also achieves the lowest average ATE.


Table \ref{tab-ETH3D} shows the comparison results on the ETH3D-SLAM benchmark with RGB-D input.
Without any finetuning on the dataset, our DVI-SLAM achieves the Top-1 place on the ETH3D-SLAM benchmark.

\begin{table}[t]
    \begin{center}
    \caption{Comparison on ETH3D-SLAM benchmark.}
    \label{tab-ETH3D}
    \resizebox{8cm}{!}
    {
        \begin{tabular}{l|c|cc|cc}\hline
            \multirowcell{2}{Method}&\multirowcell{2}{Input} & \multicolumn{2}{c|}{Max error=2cm} & \multicolumn{2}{c}{Max error=8cm} \\
             &  &AUC(train)$\uparrow$   &AUC(test)$\uparrow$ &AUC(train)$\uparrow$   &AUC(test)$\uparrow$\\ \hline
            BundleFusion\cite{dai2017bundlefusion} &RGB-D &8.29  &2.59  &84.10  &33.84 \\
            ElasticFusion\cite{whelan2015elasticfusion} &RGB-D &9.25  &2.69  &89.06 &34.02 \\
            DVO-SLAM \cite{kerl2013dense} &RGB-D &26.88  &5.83   &193.89 &71.83 \\
            ORB-SLAM3\cite{campos2021orb} &RGB-D &9.46  &5.82 &113.55 &73.33 \\
            ORB-SLAM2\cite{mur2017orb} &RGB-D &23.03 &17.84 &156.10  &104.28 \\
            BAD-SLAM\cite{schops2019bad} &RGB-D  &56.24 &30.89 &280.05 &153.47 \\
            GO-SLAM\cite{zhang2023go} &RGB-D  &- &33.12 &-  &197.02\\
            DROID-SLAM\cite{teed2021droid} &RGB-D  &54.61 &32.84   &340.42  &207.79\\ \hline
            DVI-SLAM(Ours) &RGB-D  &\textbf{62.52} &\textbf{37.97} &\textbf{354.71} &\textbf{211.60}\\ \hline
        \end{tabular}
    }
    \end{center}
    \end{table}

\subsection{Ablation Study}

Table \ref{tab_1} shows the effectiveness of our dual visual factor design that improves performance on the synthetic TartanAir validation dataset \cite{wang2020tartanair}, which is defined by DROID-SLAM \cite{teed2021droid}.
From the table, we notice that our DVI-SLAM method achieves about 40$\%$ trajectory error reduction in comparison to the baseline method on the output of VO or SLAM regardless of the monocular or stereo input.
In particular, we add feature-metric factor with different weights.
Firstly, a fixed weight is used and is named ``B + F$\_$Metric$\_$fixed $\boldsymbol{w}^f=1$", but the results become worse.
Then we manually tune the weight from small to large in the iteration process and named it ``B + F$\_$Metric$\_$tuned $\boldsymbol{w}^f$".
In comparison, our network learns better weight to dynamically fuse the dual visual factors.

\begin{table}[h]
    \begin{center}
    \caption{Ablation study of the influence of each factor on TartanAir validation dataset.}
    \label{tab_1}
    \resizebox{8cm}{!}{
    \begin{tabular}{l|l|l|l|l}\hline
    \multirowcell{2}{Configuration}    &\multicolumn{2}{c|}{\makecell{Mono\\ATE-scale(m)$\downarrow$}}  &\multicolumn{2}{c}{\makecell{Stereo\\ATE(m)$\downarrow$}} \\
      &\makecell{VO} &\makecell{SLAM} &\makecell{VO} &\makecell{SLAM} \\ \hline
    { Baseline(B)}  &0.264 &0.129 &0.100 &0.060 \\
    { B + F-Metric$\_$fixed $\boldsymbol{w}^f=1$}  &23.78 &23.98 &14.58 &14.59 \\
    { B + F-Metric$\_$tuned $\boldsymbol{w}^f$}  &0.190 &0.115 &0.078 &0.048 \\
    { B + F-Metric$\_$learned $\boldsymbol{w}^f$}  &\textbf{0.137}{\color{red}$^{48.1\% \downarrow}$} &\textbf{0.080}{\color{red}$^{38.0\% \downarrow}$} &\textbf{0.069}{\color{red}$^{31.0\% \downarrow}$} &\textbf{0.036}{\color{red}$^{40.0\% \downarrow}$} \\ \hline
    \end{tabular}
    }
    \end{center}
    \end{table}

\begin{figure}[h]
    \begin{center}
    \includegraphics[width=0.9\linewidth]{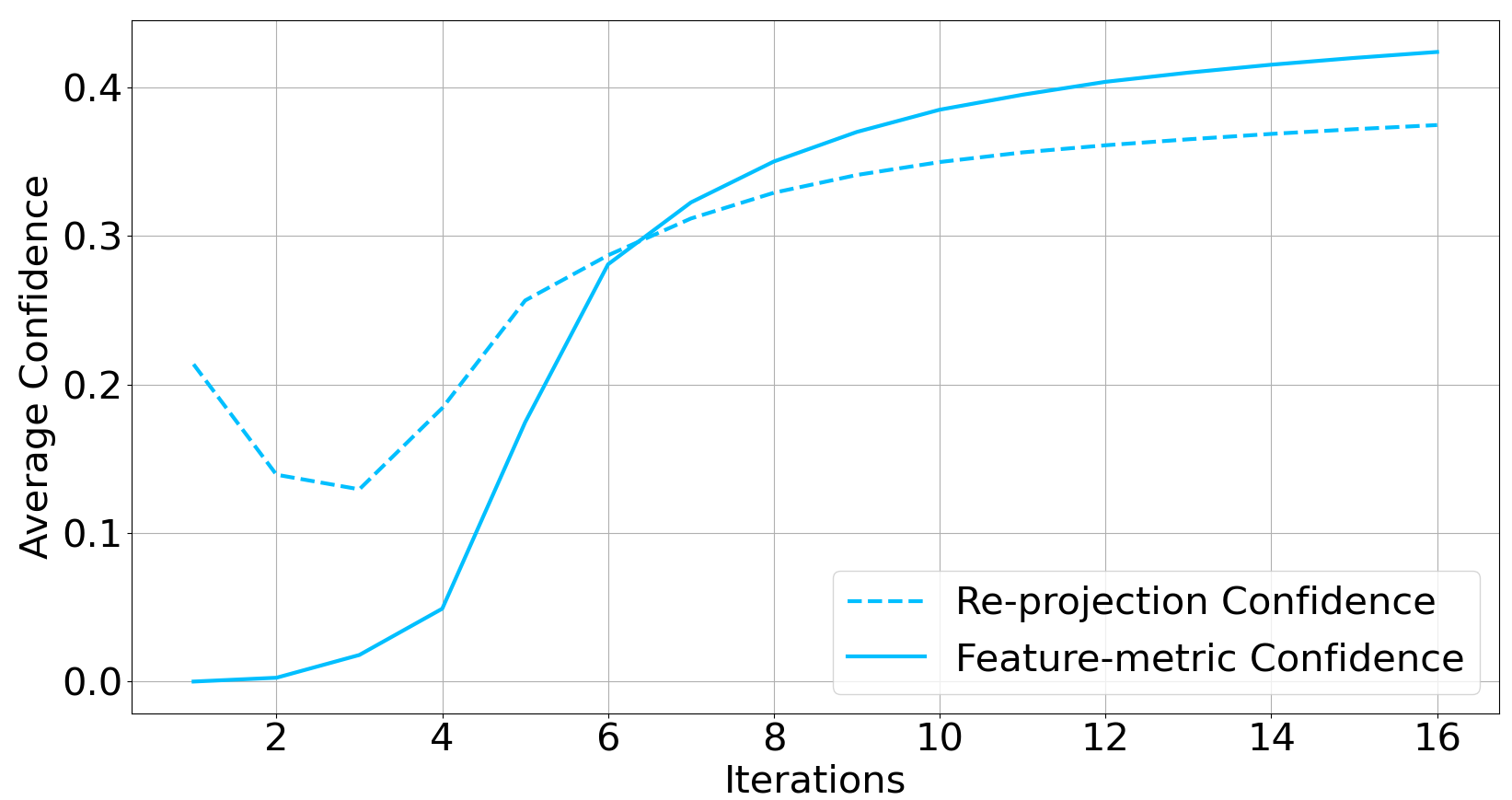}
    \end{center}
    \caption{The graph shows the change of average re-projection confidence and average feature-metric confidence with the number of iterations increasing.}
    \label{fig-weight}
    \end{figure}

Figure \ref{fig-weight} visualizes the average dual visual confidence variation with respect to the optimization iteration step when inference.
We notice that the re-projection factor dominates at the early stage. When the pose and depth state approximate correct values, feature-metric confidence increases rapidly. After several iterations, the average confidence of feature-metric exceeds the re-projection factor a little bit and they jointly contribute to the optimization.

Table \ref{tab_ablation_IMU} shows the impact of each factor on performance on the real-world EuRoC dataset.
From the table, we observe that IMU factor has a larger impact on decreasing the trajectory error in comparison with the feature-metric factor.
Specifically, we make experiments with different IMU weight.
For example, for the method named ``B + IMU$\_$fixed $\boldsymbol{w}^{u}=1$" , we only use the pre-integration weight $\boldsymbol{w}^{pu}$.
For the method named ``B + IMU$\_$tuned $\boldsymbol{w}^{u}$" , we multiply a manually tuned vector on the pre-integration weight $\boldsymbol{w}^{pu}$.
The results of this method approaches our learned performance.
When dynamically combining the three factors together, the ATE of monocular and stereo VIO has reduced by 45.3$\%$ and 36.2$\%$ respectively.

\begin{table}[h]
\begin{center}
\caption{Ablation study of the influence of each factor on the EuRoC dataset. }
\label{tab_ablation_IMU}
\resizebox{7.5cm}{!}{
 \begin{tabular}{l|l|l}\hline
  \makecell{Configuration} &\makecell{Mono(VO/VIO)\\ATE-scale(m)$\downarrow$} &\makecell{Stereo(VO/VIO)\\ATE(m)$\downarrow$}  \\ \hline
 { Baseline(B)} &0.179 &0.080 \\ \hline
 { B + F-Metric$\_$fixed $\boldsymbol{w}^f=1$} &3.203 &3.409 \\
 { B + F-Metric$\_$tuned $\boldsymbol{w}^f$} &0.163 &0.073 \\
 { B + F-Metric$\_$learned $\boldsymbol{w}^f$} &0.143{\color{red}$^{20.1\% \downarrow}$} &0.066{\color{red}$^{17.5\% \downarrow}$} \\ \hline
 { B + IMU$\_$fixed $\boldsymbol{w}^{u}=1$} &0.251 &0.127 \\
 { B + IMU$\_$tuned $\boldsymbol{w}^{u}$} &0.129 &0.064 \\
 { B + IMU$\_$learned $\boldsymbol{w}^{u}$} &0.114{\color{red}$^{36.3\% \downarrow}$} &0.062{\color{red}$^{22.5\% \downarrow}$} \\ \hline
 \makecell{B + F-Metric$\_$learned $\boldsymbol{w}^f$ \ \qquad\\+ IMU$\_$learned $\boldsymbol{w}^{u}$\ \quad \qquad} &{\textbf{0.098}}{\color{red}$^{45.3\% \downarrow}$} &{\textbf{0.051}}{\color{red}$^{36.2\% \downarrow}$} \\ \hline
\end{tabular}
}
\end{center}
\end{table}

\subsection{Memory and Runtime}
Table \ref{tab-memory} shows the memory requirements of our VIO or VI-SLAM and runtime on different datasets with different resolution inputs. All tests are implemented with one NVIDIA RTX 3090 GPU.
The runtime is the total runtime of VO or VIO divided by the total frame number of the sequence. In the table, we show the frames per second (fps), which is the inverse of the runtime.

\begin{table}[h]
    \begin{center}
    \caption{The memory requirement and runtime.}
    \label{tab-memory}
    \resizebox{8cm}{!}{
    \begin{tabular}{l|c|cc|c}\hline
        \multirowcell{2}{Dataset} &\multirowcell{2}{Resolution} &\multicolumn{2}{c|}{Memory} &\multirowcell{2}{Runtime} \\
        & &VO/VIO &V/VI-SLAM & \\ \hline
        TartanAir\cite{wang2020tartanair} &$384 \times 512$  &12(G) &24(G)    &8(fps)  \\
        EuRoC\cite{burri2016euroc}     &$240 \times 384$  &8(G)  &24(G)    &20(fps)  \\
        ETH3D-SLAM\cite{schops2019bad}     &$312 \times 504$  &12(G) &24(G)    &16(fps)  \\ \hline
    \end{tabular}
    }
    \end{center}
    \end{table}

\section{Conclusions and Future Work}

We present DVI-SLAM, a dual visual and inertial SLAM method that leverages dynamic multiple factors fusion to improve the SLAM accuracy.
We verify that our DVI-SLAM method outperforms existing SLAM methods significantly in localization accuracy and robustness through extensive experiments.
In the future, We would like to speed up the algorithm and reduce model memory requirements, for example by adjusting from dense to sparse tracking, which is an important step towards achieving real-time pose estimation on consumer-grade mobile processor platforms. In addition, in order to provide the accurate and complete 3D map, we also would like to better integrate RGB and depth sensors with Neural Radiance Fields (NeRFs) or 3D Gaussian Splatting methods.

\bibliographystyle{IEEEtran}
\bibliography{root}

\end{document}